\def\year{2020}\relax
\documentclass[letterpaper]{article} 
\usepackage{aaai20}  
\usepackage{times}  
\usepackage{helvet} 
\usepackage{courier}  
\usepackage[hyphens]{url}  
\usepackage{graphicx} 
\usepackage{graphicx}  
\usepackage{amsfonts}
\usepackage{capt-of}
\usepackage{booktabs}
\usepackage{soul,xcolor}
\usepackage{multirow}
\usepackage{nicefrac}
\usepackage{amsmath}
\usepackage[ruled]{algorithm2e}
\usepackage{enumitem}
\newcommand{\citet}[1]{\citeauthor{#1}\shortcite{#1}} \newcommand{\citep}{\cite}  

\title{Adversarial Orthogonal Regression: Two non-Linear Regressions for Causal Inference}

\author{M. Reza Heydari,\textsuperscript{\rm 1} Saber Salehkaleybar,\textsuperscript{\rm 1} Kun Zhang\textsuperscript{\rm 2}\\
\textsuperscript{\rm 1}Department of Electrical Engineering, Sharif University of Technology, heydari\_mr@ee.sharif.edu, saleh@sharif.edu\\
\textsuperscript{\rm 2}Department of Philosophy, Carnegie Mellon University, kunz1@cmu.edu}
 
\begin{document}
\setstcolor{red}
\maketitle

\begin{abstract}
We propose two nonlinear regression methods, named Adversarial Orthogonal Regression (AdOR) for additive noise models and Adversarial Orthogonal Structural Equation Model (AdOSE) for the general case of structural equation models. Both methods try to make the residual of regression independent from regressors, while putting no assumption on noise distribution. In both methods, two adversarial networks are trained simultaneously where a regression network outputs predictions and a loss network that estimates mutual information (in AdOR) and KL-divergence (in AdOSE). These methods can be formulated as a minimax two-player game; at equilibrium, AdOR finds a deterministic map between inputs and output and estimates mutual information between residual and inputs, while AdOSE estimates a conditional probability distribution of output given inputs. The proposed methods can be used as  subroutines to address several learning problems in causality, such as causal direction determination (or more generally, causal structure learning) and causal model estimation. Synthetic and real-world experiments demonstrate that the proposed methods have remarkable performance with respect to previous solutions. 
\end{abstract}

\section{Introduction}
\label{intro}

Identifying cause-effect relationships between variables in complex high dimensional networks has been studied in many fields such as neuroscience \citep{jazayeri2017navigating,shadlen1996computational}, computational genomics \citep{marbach2012wisdom,haury2012tigress}, economics \citep{zellner1988causality}, and social networks \citep{ver2012information,ver2013information}. For instance, in genomics, it is known that each cell of living creatures consists of a huge number of genes that produce proteins in a procedure called ``gene expression,'' in which they can inhibit or promote each others' activities. These cause-effect relationships can be represented by a causal graph in which each variable is depicted by a node, and a directed edge that shows the direct causal effect from the ``parent'' node to the ``child'' node. It is commonly assumed that there is no directed cycle in the causal graph, i.e., it is a Directed Acyclic Graph (DAG).  The goal is to recover the causal graph from the data sampled from variables.
In the literature, learning causal graphs has been studied extensively in two main settings: random variables and time series. 


In the setting of random variables, \citet{shimizu2006linear} proposed LiNGAM algorithm which can identify the causal graph in linear model under the assumption of non-Gaussianity of exogenous noises in the system. \citet{hoyer2009nonlinear} proposed a method to reveal the direction of causality in additive noise model where the effect is a function of direct causes plus some exogenous noise. The basic idea of their method is the following: for a given candidate DAG, one solves a regression problem for each node, modeling it as a (possibly nonlinear) function of its parents. Then, a statistical independence test is performed to assess whether all residuals are jointly independent. If that is the case, the candidate DAG is accepted, otherwise it is rejected. \citet{peters2013causal} extended this idea for time series in additive noise models. All these methods require nonparametric nonlinear regression such that it ensures the residual is independent of regressors.

%

In the setting of time series, much efforts exerted to define statistical definition of causality such as Granger causality \citep{granger1969investigating,granger1963economic}. \citet{marko1973bidirectional} defined an information theoretic measure called Directed Information (DI), which is a statistical criterion to detect the existence of direct causal effect between any pair of time series. Based on DI and inspired by G-causality, \citet{quinn2015directed} proved that minimal generative model, i.e., a graph with minimum number of edges that does not miss the full dynamics, can be discovered by causally conditioned DI. Experiments showed that the proposed criterion can be used to reconstruct efficiently the causal graphs with linear relationships.

Causally conditioned DI and the other information theoretic measures for causality in time series typically utilize ``differential entropy,'' \citep{peters2017elements} which is an extension of Shannon entropy for continuous random variables. Since differential entropy is defined based on the probability distribution, numerous works have been done for entropy estimation of general distributions using only observational data. In this regard, \citet{hausser2009entropy} used a naive binning method to estimate the value of joint distribution in each bin and then adjusted these values by a shrinkage factor based on James-Stein estimator \citep{james1992estimation}. In \citep{liu2009high,darbellay2000independent,miller2003new}, the joint distribution is estimated by partitioning the domain in such a way that more accurate values are achieved in the regions where the density of sampled data is high. However, the proposed methods are sophisticated and need huge computational cost in high dimension. 
 Recently, \citet{quinn2015directed} used a regression based method for estimating DI. In order to check whether a variable \(Y\) is the parent of variable \(X\), two regressions are performed: one by considering the \(Y\) in the regressors, and another without it. Then, DI can be obtained by differing the entropy of residuals in two regressions. The \(Y\) is considered as a parent of \(X\) if DI is non-zero. The above procedure works correctly only if the obtained residuals are independent of regressors in both regressions.


According to what mentioned above, several causal learning algorithms in the setting of random variables (such as the one in \citep{hoyer2009nonlinear}) or time series (such as TiMINo algorithm in \citep{peters2013causal} or DI estimator in \citep{quinn2015directed}),  require a subroutine that can perform non-linear regression such that the residual becomes independent of the regressors as much as possible.
However, common regression methods are confined to minimize Mean Squared Error (MSE) loss \citep{wang2009mean,kay1993fundamentals}. Thus, in these common methods, the residuals and regressors become only uncorrelated. While these methods are fully efficient in linear Gaussian case, they might not be statistically efficient in nonlinear or non-Gaussian scenarios. To resolve this issue, \citet{mooij2009regression} proposed a novel regression method which minimizes the dependence between residuals and regressors that is meausured by Hilbert-Schmidt Independence Criterion (HSIC).
In the proposed method, it is needed to carefully tune the kernel parameter in HSIC.

\textbf{Contributions:} In this paper, we propose two nonlinear regression methods, named Adversarial Orthogonal Regression (AdOR) and Adversarial Orthogonal Structural Equation Model (AdOSE). AdOR assumes that the noise is modeled as an additive term while AdOSE relaxes this assumption. The models are ``Adversarial'', in the sense that in both methods, two neural networks compete with each other, the regression network and the loss network. In AdOR, the loss network estimates the mutual information between regressors and residuals, and in AdOSE, it acts as a Kullback-Leibler (KL)-divergence estimator between correct responses and predicts (which are the output of regression network). As discussed above, independence of residuals and regressors is vital in inferring the correct causal relationships. Thus, AdOR tries to make the residual  independent of regressors, and AdOSE achieves this target by independently generating noise. The proposed methods can be used as  subroutines to address several learning problems in causality, such as determining causal direction, causal structure learning, or causal model estimation. Experiment show that the proposed methods have remarkable performance in estimating the true non-linear function with respect to previous solutions.
While our main contribution is in causal inference, the proposed methods might also be useful in the other regression tasks.

The rest of the paper is organized as follows: In Section \ref{mi_intro}, we review a neural network \citep{belghazi2018mine} that has been proposed previously to estimate mutual information. We describe AdOR and AdOSE methods in Section \ref{ador} and Section \ref{adose}, respectively. We provide experimental results in Section \ref{exps} and conculde the paper in Section \ref{con}.

\section{Mutual Information Neural Estimation}
\label{mi_intro}
In this section, we describe the neural network proposed in \citep{belghazi2018mine} for estimating mutual information based on an alternative representations of KL-divergence. This representation will be exerted as the loss network in Section \ref{ador} and \ref{adose}.

Let \(P\) and \(Q\) be  two distributions on some compact domain \(\Omega\subset\mathbb{R}^d\). The KL-divergence between them is defined as:
\begin{gather}\label{eq:kl_def}
D_{KL}\left(P\middle|\middle|Q\right) := \mathbb{E}_{P}\left[\text{log} \frac{dP}{dQ}\right].
\end{gather}
One of the representation of KL-divergence, which we focused on, is Donsker-Varadhan representation \citep{donsker1983asymptotic}:
\begin{gather}\label{eq:DV_rep}
D_{KL}\left(P\middle|\middle|Q\right) = \sup_{T:\Omega \to \mathbb{R}} {\mathbb{E}_{P}\left[T\right]-\text{log}\left(\mathbb{E}_{Q}\left[e^T\right]\right)},
\end{gather}
where the supremum is taken over all functions \(T\) such that the two expectations are finite.

 Let \(X\) and \(Y\) denote two continuous random variables with distributions \(P_X\) and \(P_Y\), respectively. Mutual information between \(X\) and \(Y\) is denoted by \(I\left(X;Y\right)\), which is a measure for the dependence of them. mutual information has some multiple forms, and one form is defined as the KL-divergence between the joint distribution \(P_{XY}\), and the product of marginal distributions \(P_{X}P_{Y}\):
\begin{gather}\label{eq:MI_KL}
I\left(X;Y\right) = D_{KL}\left(P_{XY}\middle|\middle|P_{X}P_{Y}\right).
\end{gather}
Let \(\mathcal{F} = \left\{T_\theta\right\}_{\theta\in\Theta}\) be the set of functions parametrized by a neural network (i.e. weights, biases, batch normalization parameters, etc.). Mutual Information Neural Estimator (MINE, see definition 3.1 of \citep{belghazi2018mine}) is defined as:
\begin{gather}\label{eq:MINE}
\hat{I}\left(X;Y\right) = \sup_{\theta\in\Theta} {\mathbb{E}_{P_{XY}}\left[T_\theta\right]-\text{log}\left(\mathbb{E}_{P_{X}P_{Y}}\left[e^{T_\theta}\right]\right)}.
\end{gather}
As the class of all functions in (\ref{eq:DV_rep}) is restricted to neural network class \(\mathcal{F}\) in (\ref{eq:MINE}), we have the following lower bound:
\begin{gather}\label{eq:lower_bound}
I\left(X;Y\right) \geq \hat{I}\left(X;Y\right).
\end{gather}
Theoretical properties of $\hat{I}\left(X;Y\right)$ are provided in \citep{belghazi2018mine}. In MINE, samples from joint distribution \(P_{XY}\) are fed as the inputs of a neural network and an optimizer like stochastic gradient descent, updates the parameters \(\theta\) so as to maximize the right hand side of (\ref{eq:MINE}). Ultimately, as the parameters converge, the loss value of network is the estimated mutual information. For more details on the implementation of MINE, please refer to Algorithm 1 of \citep{belghazi2018mine}.

\section{Adversarial Orthogonal Regression}
\label{ador}
 Let \(Z\) and \(U\) represent the scalar response and regressor vector, respectively. The regression problem is to find \(\hat{f}\):
\begin{gather}\label{eq:Reg_add}
Z = \hat{f} \left(U\right) + \varepsilon,
\end{gather}
such that the residual \(\varepsilon\) is independent of \(U\).

In AdOR method, the regression network (\(R\)) is pitted against the loss network where a mutual information estimator (\(MI\)) learns to find any high order dependencies (see the top block diagram of Figure \ref{fig:AdOR_AdOSE_Sc}). In regression part, $\hat{Z}=\hat{f}\left(U;\theta_R\right)$  is a differentiable function represented by a multilayer perceptron, and parametrized with \(\theta_R\), in which \(\hat{Z}\) is the regression output. The residual \(\varepsilon=Z-\hat{Z}\) and the regressor vector \(U\) are fed as inputs to \(MI\), and the output \(T\left(\varepsilon,U;\theta_{MI}\right)\) is also a differentiable function represented by a multilayer perceptron with parameters \(\theta_{MI}\).  $L\left(R,MI\right)=\mathbb{E}_{P_{\varepsilon U}}\left[T\right]-\text{log}\left(\mathbb{E}_{P_\varepsilon P_U}\left[e^T\right]\right)$ denotes the mutual information between $U$ and $\varepsilon$. \(R\) is trained to minimize the dependency between residual and regressors. \(MI\) is simultaneously trained to tighten the gap between  $I\left(U;\varepsilon\right)$ and  $\hat{I}\left(U;\varepsilon\right)$ in order to achieve more accurate estimate of mutual information. In other words, \(R\) and \(MI\) play the following two-player minimax game:
\begin{gather}\label{eq:MiniMax}
\min_{R}{\max_{MI}{L\left(R,MI\right)=\mathbb{E}_{P_{\varepsilon U}}\left[T\right]-\text{log}\left(\mathbb{E}_{P_\varepsilon P_U}\left[e^T\right]\right)}}
\end{gather}

At equilibrium point, the value of loss \(L\left(R,MI\right)\) is mutual information between \(U\) and \(\varepsilon\). We provide experimental results in Section \ref{imp_deta} that show convergence to the equilibrium point. In practice, the game in (\ref{eq:MiniMax}) is implemented by an iterative approach, in which the gradient of loss \(\nabla L_{B}\) for mini-batch \(B\) is used via back-propagation procedure. As mentioned in \citep{belghazi2018mine}, the second term in the mini-batch's gradient \(\nabla L_{B}\) leads to a biased estimate of the full-batch gradient \(\nabla L\). To overcome this issue, Adam optimizer \citep{kingma2014adam} can be utilized where the history of gradients is also considered in the next update.

  \begin{algorithm}
    \caption{AdOR}\label{alg:AdOR}
    \For{number of iterations}
    	{\textbf{Forward path}:
    	\begin{enumerate}[leftmargin=*]
    	\item Draw \(2b\) minibatch samples \(\left\{\left(u^{(1)},z^{(1)}\right),\dotsc,\left(u^{(2b)},z^{(2b)}\right)\right\}\)
    	\item Evaluate regression output 
    	
    	\(\hat{z}^{(i)}=\hat{f}\left(u^{(i)};\theta_R\right); i=1,\dotsc,2b\)
    	\item Compute residual  
    	
    	\(\varepsilon^{(i)} = z^{(i)} - \hat{z}^{(i)}; i=1,\dotsc,2b\)
    	\item Evaluate output of \(MI\) twice \(T^{(i)}=T\left(\varepsilon^{(i)},u^{(i)};\theta_{MI}\right); i=1,\dotsc,b\)
    	\(T_{sh}^{(i)}=T\left(\varepsilon^{(i+b)},u^{(i)};\theta_{MI}\right); i=1,\dotsc,b\)
    	\item Compute loss 
    	\small
    	\(L_B\left(\theta_R,\theta_{MI}\right)=\frac{1}{b}\sum_{i=1}^{b}{T^{(i)}}-\text{log}\left(\frac{1}{b}\sum_{i=1}^{b}{e^{T_{sh}^{(i)}}}\right)\)
    	\end{enumerate}
    	
    	\textbf{Backward path}:
    	
    	\For{\(k_R\) steps}
    	{\small Update \(R\) by descending its stochastic gradient \(\nabla_{\theta_R} L_B\)}
    	
    	\For{\(k_{MI}\) steps}
    	{\small Update \(MI\) by ascending its stochastic gradient \(\nabla_{\theta_{MI}} L_B\)}
    	}
  \end{algorithm}

Algorithm \ref{alg:AdOR} shows AdOR training. In forward path, \(2b\) examples are fed to \(R\), and residuals \(\varepsilon^{(i)}\) are computed in line 3. The first \(b\) pairs \(\varepsilon^{(i)}\) and \(u^{(i)}\) are jointly sampled; while, the second \(b\) pairs \(\varepsilon^{(i+b)}\) and \(u^{(i)}\) are marginal samples. Output of \(MI\) is computed twice: once by joint samples, and once by marginal samples in line 4. Finally, mini-batch loss \(L_B\) is computed in line 5 based on mean of samples computed in line 4. In backward path, parameters of each network are updated while the ones of other network is fixed. Note that in each iteration, \(R\) and \(MI\) are updated \(k_R\) and \(k_{MI}\) times, respectively. 

\section{Adversarial Orthogonal Structural Equation Model}
\label{adose}

In \eqref{eq:Reg_add}, the noise \(\varepsilon\) is modeled as an additive term. However, in general, the exogenous noise can affect the variable \(Z\) in a non-linear form, such as in structural equation models (SEM, see \citep{peters2017elements}). Thus, we assume here that the true model is: \(Z=f(U,\varepsilon)\). In AdOSE, we propose a new method to estimate both the nonlinear function \(f\) and also the joint distribution \(P_{UZ}\). Hence, our goal is to obtain a function \(\hat{f}\):
\begin{gather}\label{eq:Reg_non_add}
\hat{Z} = \hat{f} \left(U,\varepsilon\right),
\end{gather}
such that \(\hat{Z}\) is similar as possible as to the response \(Z\), with the same \(U\); i.e. \(D_{KL}\left(P_{UZ}\middle|\middle|P_{U\hat{Z}}\right)=0\).

\begin{figure*}[t]
	\begin{center}
		\includegraphics[width=0.95\linewidth]{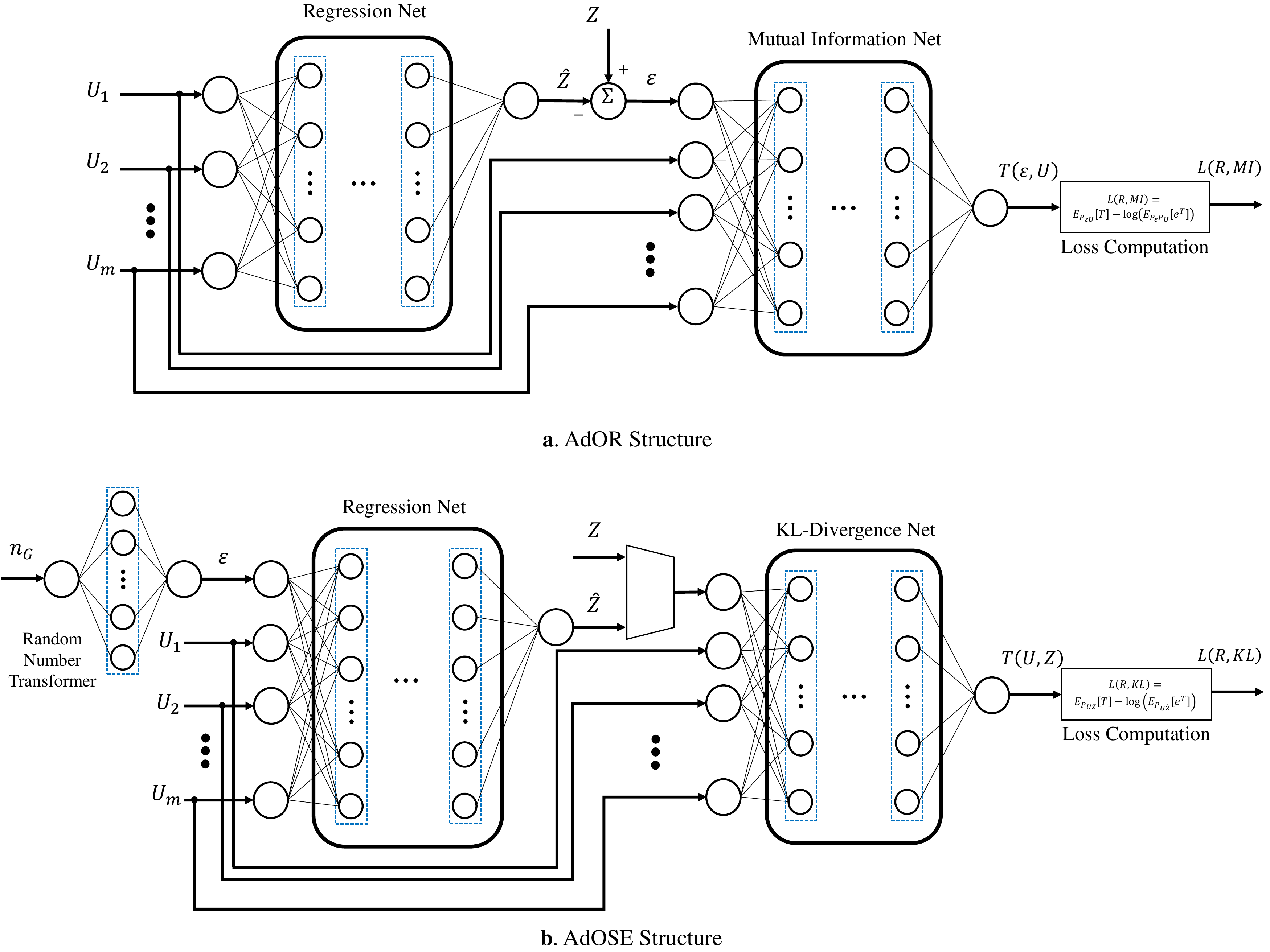}
		\caption{\label{fig:AdOR_AdOSE_Sc}Block diagram of AdOR and AdOSE. \textbf{a}. AdOR structure: \(\left(U_1,\dotsc,U_m\right)\) are the input regressors, \(\hat{Z}\) is the predict, and \(\varepsilon\) is the residual. The output loss \(L\left(R,MI\right)\) is the estimated mutual information between them. \textbf{b}. AdOSE structure: \(n_G\) is generated by Gaussian generator, \(\varepsilon\) is the exogenous noise, \(\left(U_1,\dotsc,U_m\right)\) and \(\varepsilon\) are fed as inputs to \(R\). \(KL\) computes the output twice: once using \(\left(U_1,\dotsc,U_m,Z\right)\), and once by \((U_1,\dotsc,U_m,\hat{Z})\). The output \(L\left(R,KL\right)\) is the KL-Divergence.}
	\end{center}
\end{figure*}

In AdOSE, similar to AdOR, the regression network (\(R\)) is pitted against the loss network: a KL-divergence estimator (\(KL\)) that learns to match the joint distribution \(P_{U\hat{Z}}\) to distribution \(P_{UZ}\) (see the bottom diagram of Figure \ref{fig:AdOR_AdOSE_Sc}). Inspired by GAN (\citet{goodfellow2014generative}), in AdOSE, the noise \(n_G\) is generated by a random Gaussian generator and transformed to the noise \(\varepsilon\) through a one-hidden layer perceptron \(RanTrans\); i.e. \(\varepsilon=RT\left(n_G\right)\). Then, regressors \(U\) and generated noise \(\varepsilon\) are passed to the regression network \(R\), similar to AdOR; $\hat{Z}=\hat{f}\left(U,\varepsilon;\theta_R\right)$. Afterwards, pairs \((U,Z)\) and \((U,\hat{Z})\) are passed through \(KL\) by a differentiable transformation \(T\), and the outputs are \(T(U,Z;\theta_{MI})\) and \(T(U,\hat{Z};\theta_{MI})\), respectively. Based on (\ref{eq:DV_rep}), the KL-distance is estimated by $L\left(R,KL\right)=\mathbb{E}_{P_{UZ}}\left[T\right]-\text{log}\left(\mathbb{E}_{P_{U\hat{Z}}}\left[e^T\right]\right)$, and two networks play the following minimax game:
\begin{gather}\label{eq:MiniMax_AdOSE}
\min_{R}{\max_{KL}{L\left(R,KL\right)=\mathbb{E}_{P_{UZ}}\left[T\right]-\text{log}\left(\mathbb{E}_{P_{U\hat{Z}}}\left[e^T\right]\right)}}.
\end{gather}

At equilibrium, the value of loss \(L\left(R,KL\right)\) is zero. After training, instead of having a nonlinear mapping between regressors and response, we have a nonlinear transformation for each samples of \(U=u\), that assigns a distribution for \(Z\); i.e. \(\hat{Z}\sim P(Z|U=u)\). Indeed, as the true value of \(n_G\) is unknown, we can not obtain single predict for each input sample \(u\); while, we can draw output samples by feeding different values of \(n_G\). Since training AdOSE is more trickier than AdOR, we provide some implementation details in Section \ref{imp_deta} to avoid divergence of the algorithm.

Algorithm \ref{alg:AdOSE} shows the training procedure of AdOSE. In forward path, \(b\) Gaussian samples are drawn and fed to \(RanTrans\). The regression output is computed in line 3. As \(MI\) in AdOR, \(KL\) evaluates \(T\) twice: once by using \(u^{(i)}\) and true responses \(z^{(i)}\), and once by \(u^{(i)}\) and predicted responses \(\hat{z}^{(i)}\) (line 4). Mini-batch loss \(L_B\) is then computed using mean of true and estimated \(T\). Similar to AdOR, in backward path, \(k_R\) and \(k_{KL}\) control the training of two networks. Furthermore, they play the main rule in convergence of the algorithm; if the loss is large, \(R\) has bad predicts and \(k_R\) should be increased, and if it is small, \(KL\) can not distinguish between true and predicted values and \(k_{KL}\) should be increased.

\subsection*{Applications in Causal Inference}
\label{CIADOR}

AdOR and AdOSE can be used in causal models that assume there is a structural model between child and parents. For instance, consider the additive noise model (ANM) between the cause variable \(C\) and the effect variable \(E\): \(E=f(C)+\varepsilon\). In \citep{hoyer2009nonlinear}, it has been shown that there exist no function \(g\) and noise \(\tilde{\varepsilon}\) almost surely such that \(C=g(E)+\tilde{\varepsilon}\) and \(E\) and \(\tilde{\varepsilon}\) are independent. Hence, we can utilize AdOR to infer causal direction between two variables \(X\) and \(Y\). To do so, we regress each variable on the other one and pick the direction with minimum loss \(L(R,MI)\).  
 Moreover, one can use AdOR as the class of functions for TiMINO (\citet{peters2013causal}) for inferring causal direction in time series. At last, the causally conditioned DI \citep{quinn2015directed} of each child on each candidate parent can also be estimated by regress the child twice, one on all variables, and the other on all variables except the candidate parent. The difference of two residuals' entropy is DI from parent to child.

  \begin{algorithm}
  	
    \caption{AdOSE}\label{alg:AdOSE}
    \For{number of iterations}
    	{\textbf{Forward path}:
    	\begin{enumerate}[leftmargin=*]
    	\item Generate \(b\) Gaussian samples \(\{n_G^{(1)},\dotsc,n_G^{(b)}\}\) 
    	
    	 Feed them to \(RanTrans\): \(\varepsilon^{(i)}=RT(n_G^{(i)})\)
    	\item Draw \(b\) minibatch examples 
    	
    	\(\left\{\left(u^{(1)},z^{(1)}\right),\dotsc,\left(u^{(b)},z^{(b)}\right)\right\}\)
    	\item Evaluate regression output \(\hat{z}^{(i)}=\hat{f}\left(\varepsilon^{(i)},u^{(i)};\theta_R\right); i=1,\dotsc,b\)
    	\item Evaluate output of \(KL\) twice \(T^{(i)}=T\left(z^{(i)},u^{(i)};\theta_{KL}\right); i=1,\dotsc,b\)
    	\(T_{es}^{(i)}=T\left(\hat{z}^{(i)},u^{(i)};\theta_{KL}\right); i=1,\dotsc,b\)
    	\item Compute loss 
    	\small
    	\(L_B\left(\theta_R,\theta_{KL}\right)=\frac{1}{b}\sum_{i=1}^{b}{T^{(i)}}-\text{log}\left(\frac{1}{b}\sum_{i=1}^{b}{e^{T_{es}^{(i)}}}\right)\)
    	\end{enumerate}
    	
    	\textbf{Backward path}:
    	
    	\For{\(k_R\) steps}
    	{\small Update \(R\) and \(RanTrans\) by descending its stochastic gradient \(\nabla_{\theta_R} L_B\)}
    	
    	\For{\(k_{KL}\) steps}
    	{\small Update \(KL\) by ascending its stochastic gradient \(\nabla_{\theta_{KL}} L_B\)}
    	}
  \end{algorithm}

\section{Experiments}
\label{exps}
In this section, we first evaluate the performance of proposed regression methods on synthetic data and compare with the method in \citep{mooij2009regression} and some other nonlinear regression methods. Then, we apply the proposed method to find the causal direction in some real-world bilinear data \citep{mooij2016distinguishing}.

\subsection{Implementation Details}
\label{imp_deta}
The main point in training both AdOR and AdOSE is that the two networks \(R\) and \(MI\) (\(KL\) in AdOSE) should be trained simultaneously. As discussed before, Adam optimizer \citep{kingma2014adam} is used, and all weights and biases initialized using Xavier initializer \citep{glorot2010understanding}. The number of layers, learning rate, and batch size are chosen similar in both networks.

In AdOR, we use three hidden layers with \(tanh\), \(sigmoid\) and \(leaky{-}ReLU\) activation functions for \(R\) and three hidden layers with \(leaky{-}ReLU\) activation for \(MI\). Note that adding a bias term to \(\hat{f}(U)\) in (\ref{eq:Reg_add}) does not change mutual information, so bias term is removed from output layer of \(R\). Similarly, adding a constant term to \(T\left(\varepsilon,U;\theta_{MI}\right)\) does not change the computed loss \(L(R,MI)\) in (\ref{eq:MiniMax}), and we omit the bias term from output layer of \(MI\). Instead, the maximum mini-batch value \(\max_{i=1,\dotsc,b}\{T^{(i)},T_{sh}^{(i)}\}\) is reduced from whole \(T^{(i)}\) and \(T_{sh}^{(i)}\) in order to obtain a stable computation of loss.

 The structure of AdOSE layers are designed similar to AdOR. The noise \(n_G\) is generated by normal Gaussian distribution, and \(RanTrans\) has a hidden layer with \(leaky{-}ReLU\) activation. The bias term is added to the output layer of \(R\), and biases in \(KL\) are similar to \(MI\). Finding the stable solution of AdOSE is more trickier than AdOR. The optimizer might diverge in the first few iterations, because one of networks \(R\) or \(KL\) outstrips the other. To avoid this, we adjust steps \(k_R\) and \(k_{KL}\) by looking at the value of loss \(L_B\) in each iteration in order to stabilize the training procedure. A simple choice of steps has a linear feedback form \(k_R = \lfloor a+bL_B \rfloor\) and \(k_{KL} = \lfloor a-bL_B \rfloor\). We used \(a=30\) and \(b=10\) in our simulations.

\subsection{Toy Examples}
\label{naive_exp}

\begin{figure*}[!ht]
  \begin{center}
  	\includegraphics[width=0.95\linewidth]{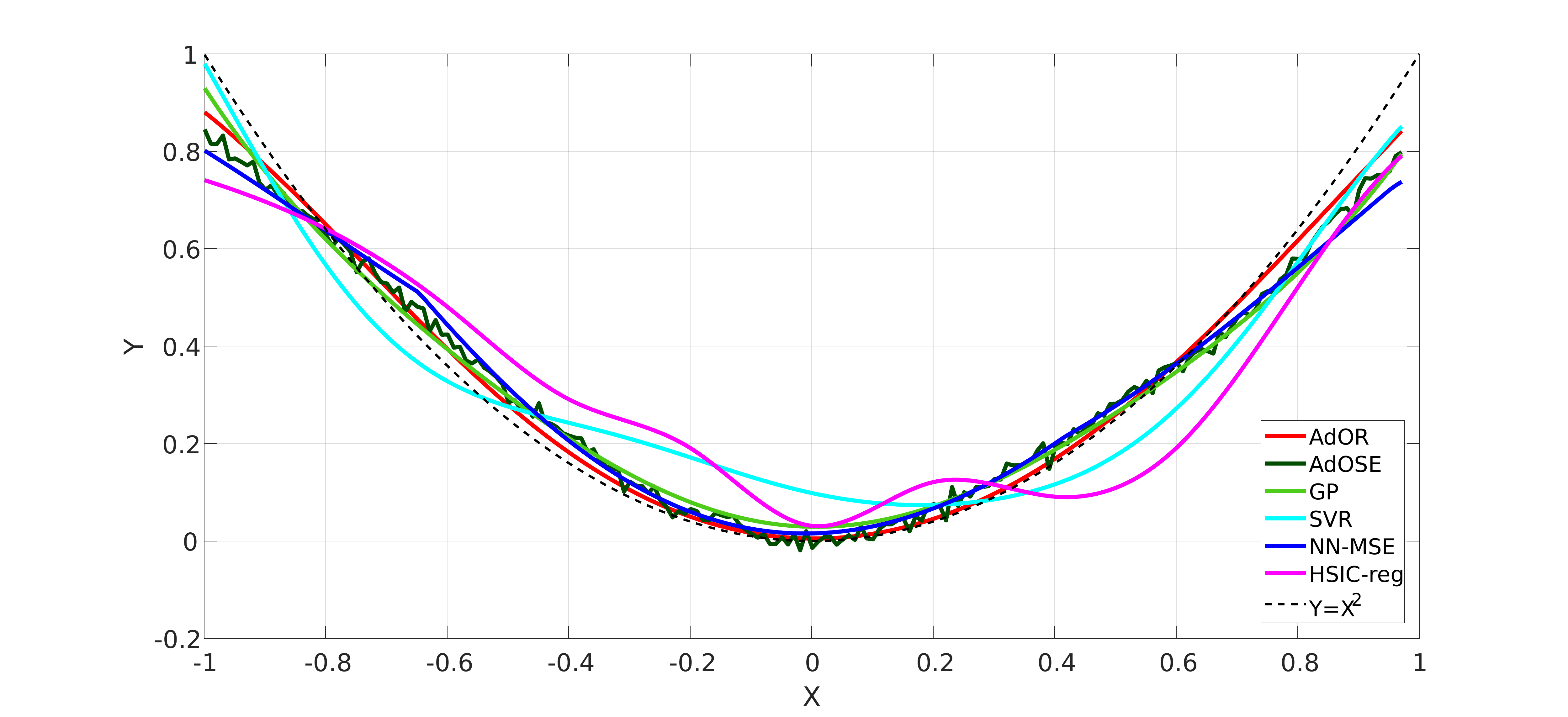}
  	\caption{\label{fig:samp_par_exp} Example of different methods' outputs. \(Y = X^2+\varepsilon\) and \(\varepsilon\sim Exp(1)\).}
  \end{center}
  
  \captionof{table}{\label{tab:ador_comp} Comparison of Different methods.}
  
  \centering
  \begin{tabular}{cccccc}
    \toprule
     \multicolumn{2}{c}{Model \(f(X)\)} & \(X^2\) & \(\sin(\pi X)\) & \(e^{2X}\) & \(sigmoid(5X)\)     \\
     \multicolumn{2}{c}{Noise} & \(\varepsilon\sim Exp(1)\) & \(\varepsilon\sim ChiSqr(3)\) & \(\varepsilon\sim Rayl(4)\) & \(\varepsilon\sim BioNom(20,0.3)\)    \\
    \midrule
     & SVR  & 8.320e-01 & 5.651e+00 & 6.320e+00 & 3.849e+00 \\
     & HSIC-reg & 8.419e-01  & 5.688e+00 & 6.386e+00 & 3.878e+00 \\
     & NN-MSE       & 8.373e-01 & 5.548e+00 & 6.226e+00 & 3.707e+00 \\
    \multirow{6}{*}[28pt]{MSE} & GP & 8.262e-01 & 5.586e+00 & 6.228e+00 & 3.846e+00 \\
     & AdOSE & 8.301e-01 & 7.658e+00 & 9.708e+00 & 5.112e+00 \\
     & AdOR & 9.299e-01 & 9.740e+00 & 1.209e+01 & 4.073e+00 \\
    \midrule
     & SVR  & 6.945e-01 & 1.926e+00 & 2.031e+00 & 1.555e+00 \\
     & HSIC-reg &  7.049e-01 & 1.933e+00 & 2.051e+00 & 1.561e+00 \\
    & NN-MSE       & 7.061e-01 & 1.909e+00 & 2.037e+00 & 1.531e+00 \\
     \multirow{6}{*}[29pt]{MAE} & GP & 6.975e-01 & 1.918e+00 & 2.035e+00 & 1.559e+00 \\
     & AdOSE       & 7.008e-01 & 2.071e+00 & 2.406e+00 & 1.527e+00\\
     & AdOR       & 7.426e-01 & 2.492e+00 & 2.768e+00 & 1.626e+00\\
     \midrule
     & SVR  & 1.160e-02 & 1.363e-01 & 2.197e-01 & 1.669e-02 \\
     & HSIC-reg & 2.666e-02 & 1.985e-01 & 2.551e-01 & 1.209e-01 \\
    & NN-MSE       & 8.430e-03 & 2.634e-01 & 1.078e-01 & 2.866e-01 \\
    \multirow{6}{*}[30pt]{ISE} & GP & 5.247e-03 & 1.422e-01 & 3.491e-02 & 2.059e-02 \\
     & AdOSE      & 5.109e-03 & 6.633e-02 & 8.289e-02 & 1.554e-02 \\
     & AdOR       & \textbf{1.734e-03} & \textbf{4.974e-02} & \textbf{1.908e-02} & \textbf{2.232e-03}\\
    \bottomrule
  \end{tabular}
\end{figure*}

In this part, AdOR and AdOSE are compared with four regression methods: Support Vector Regression \citep{smola2004tutorial}, neural network with same structure as AdOR with MSE loss minimization, HSIC regression proposed by \citet{mooij2009regression}, and Gaussian Process regression \citep{williams1996gaussian} with RBF kernel. The model has a simple form of \(Y=f(X)+\varepsilon\). In each test, 300 samples are drawn from uniform distribution \(X \sim U(-1,1)\). The function \(f(.)\) is nonlinear and \(\varepsilon\) is generated from different non-Gaussian distributions. Note that for AdOSE, the averaged \(\mathbb{E}_\varepsilon \left[Y|X=x\right]\) is plotted by feeding 5000 samples of \(n_G\) at each \(X=x\). Figure \ref{fig:samp_par_exp} shows the output of different methods for the case of \(f(X)=X^2\) and \(\varepsilon\sim Exponential(1)\). 

Comparison between methods is shown in Table \ref{tab:ador_comp} for different performance measures of Mean Squared Error (MSE), Mean Absolute Error (MAE) between predictions and responses, and Integral Squared Error (ISE) between estimated function and $f(x)$. As can be seen, in each case, AdOR has the worst MSE and MAE among the others; in contrast, its performance is much better in terms of ISE measure. 
In fact, we expect that AdOR/AdOSE  do not have  better performance in terms of MSE/MAE, compared to regression methods minimizing squared losses (or similar losses) since the goal of such methods is actually to minimize MSE while AdOR tries to minimize the mutual information between the residual and regressors. Moreover, it is not guaranteed that regression methods with square loss error estimate the underlying function statistically efficiently in cases other than Gaussian additive noise. In such cases, although the squared loss is minimized, the result might be dependent on the regressors. For instance, in Figure 2, in which the additive noise has a exponential distribution, it can be seen that AdOR finds the best approximation of the true function while the estimates given by NN-MSE and SVR are not close enough to it.


\subsection{Distribution Estimation with AdOSE}
\label{dist_est}

\begin{figure*}[!ht]
\begin{center}
  	\includegraphics[width=1\linewidth]{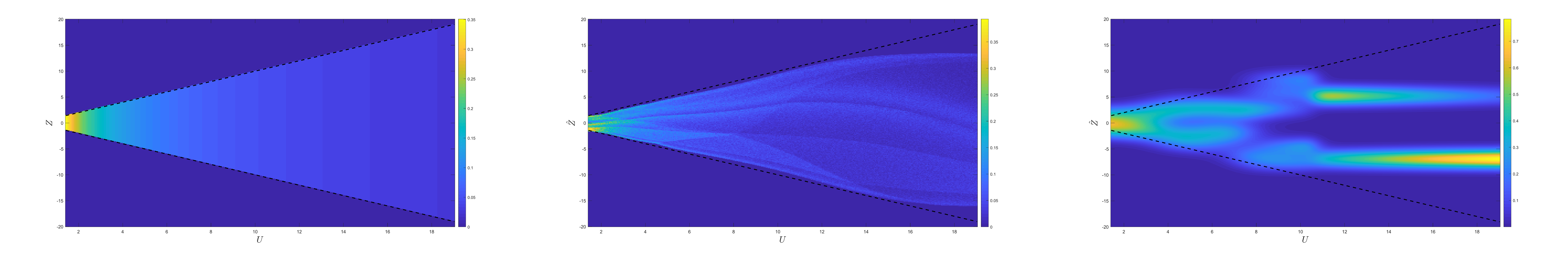}
  	\caption{\label{fig:Adose_Dist} \textbf{Left}: True conditional distribution \(P(Z|U)\). \textbf{Center}: Estimated conditional distribution \(P(\hat{Z}|U)\) by AdOSE. \textbf{Right}: Estimated conditional distribution \(P(\hat{Z}|U)\) by \citep{sugiyama2010least}.}
  \end{center}
  
  \begin{center}
  	\includegraphics[width=0.95\linewidth]{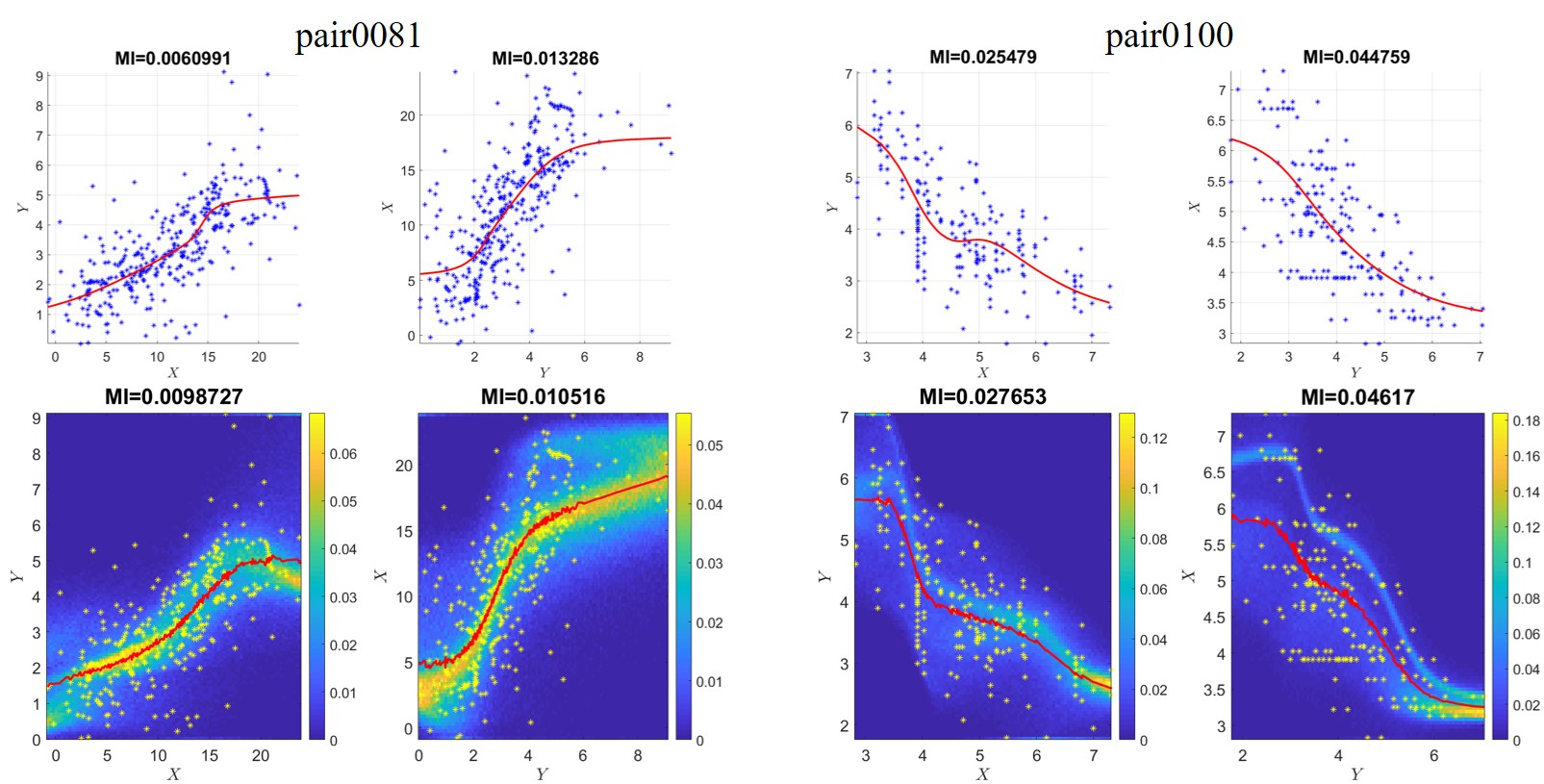}
  	\caption{\label{fig:CE_results} Results of proposed methods on real datasets. \textbf{Top row}: AdOR \textbf{Bottom row}: AdOSE}
  \end{center}
\end{figure*}

Now suppose a non-additive model \(Z=\varepsilon U\) with \(U\sim logNormal(1,0.6)+1\) and \(\varepsilon\sim Uniform(-1,+1)\).  \(1000\) number of samples are drawn from this model, and AdOSE is trained by these samples. Afterwards, \(10^5\) samples are drawn from the learned model by feeding different noise \(n_G\) for each value of \(U=u\). The conditional distribution \(P(\hat{Z}|U=u)\) is then estimated by naive binning for each \(u\) in the valid range. We also trained the model proposed by \citet{sugiyama2010least}. True conditional distribution \(P(Z|U=u)\) is depicted versus two estimated distributions \(P(\hat{Z}|U=u)\) in Figure \ref{fig:Adose_Dist}. As can be seen, the AdOSE has a great capacity to model distributions even in the regions with few samples.

\subsection{Causal Direction Discovery in Real-World Datasets}
\label{cause_effect_pairs}
\verb+Cause-effect+ (version 1.0) pairs \citep{mooij2016distinguishing} is a collection of \(108\) real-world datasets, each with different sample size from \(94\) to \(16382\), where we considered \(94\) number of these datasets. Each dataset consists of samples of two statistically dependent random variables \(X\) and \(Y\), where one variable is known to causally influence the other. The task is to infer which variable is the cause and which one is the effect.

AdOR is trained with each dataset twice: once when \(Y\) is response and \(X\) is regressor and once in the reverse direction. The direction with lower mutual information is considered as the true direction. Experimental results show that we can infer the true direction for \(66/94\) \((70.2\%)\) fraction of datasets. We defined the score \(S_i = MI(\varepsilon_{Y \to X},Y)-MI(\varepsilon_{X \to Y},X)\) for each dataset \(i\). AdOR has AUPR (Area Under Precision-Recall Curve) of \(78.68\) based on the scores \(S_i\) and its performance is similar to the best AUPR achieved by the previous methods considered in \citep{goudet2017causal}.
In the same manner, AdOSE is trained twice in forward and reverse directions. The estimated function \(\hat{f}(x) = \mathbb{E}_\varepsilon \left[Y|X=x\right]\) is computed by feeding \(5000\) samples of \(n_G\) at each \(X=x\). AdOSE can infer the true direction for \(63/94\) (\(67.0\%\)) fraction of datasets with AUPR of \(74.32\) based on the scores \(S_i\). Figure \ref{fig:CE_results} shows the estimated functions of AdOR and AdOSE on two pairs \verb+pair0081+ and \verb+pair0100+. The results for other datasets are given in the supplementary material.

\section{Conclusions}
\label{con}

We introduced two novel regression methods: AdOR which minimizes mutual information between the residual and the regressors, and AdOSE which produce response that mimics the true output by reducing distance between joint distributions. Conducted with details, we implemented our methods through adversarial neural networks and showed their great potential for inferring causal influences in models with unknown noise distributions. As a future work, one can extend these methods to the cases with categorical variables or utilize them in other causal learning problems such as learning causal structures.

\small
\bibliographystyle{aaai}\bibliography{refs}

\end{document}